%% file: main.tex
\documentclass[10pt,twocolumn,letterpaper]{article}

\usepackage[pagenumbers]{cvpr} 
\usepackage{xcolor}
\definecolor{cvprblue}{rgb}{0.21,0.49,0.74}
\definecolor{lakeblue}{rgb}{0.000000,0.650980,0.784314}
\usepackage[pagebackref,breaklinks,colorlinks,allcolors=cvprblue]{hyperref}
\usepackage{url}
\usepackage{amssymb}
\usepackage{algorithm}
\usepackage{algpseudocode}
\usepackage{booktabs}
\usepackage{multirow}
\usepackage{graphicx}
\usepackage{listings}
\usepackage{makecell}
\def\confName{CVPR}
\def\confYear{2027}

\algrenewcommand{\algorithmiccomment}[1]{
\Statex
  {\raggedleft
   \textcolor{lakeblue}{$\triangleright$\nobreakspace #1}\par}
}

\newfloat{listing}{t}{lop}
\floatname{listing}{Listing}
\definecolor{codebg}{RGB}{248,248,248}
\definecolor{codestring}{RGB}{0,110,90}
\definecolor{codekeyword}{RGB}{30,70,160}
\definecolor{codeframe}{RGB}{210,210,210}

\lstdefinelanguage{json}{
  basicstyle=\ttfamily\small,
  morestring=[b]",
  stringstyle=\color{codestring},
  morekeywords={true,false,null},
  keywordstyle=\color{codekeyword}\bfseries,
  sensitive=true,
  showstringspaces=false,
  breaklines=true,
  breakatwhitespace=false,
  columns=fullflexible,
  keepspaces=true,
  numbers=left,
  numberstyle=\scriptsize\color{gray},
  numbersep=8pt,
  frame=single,
  rulecolor=\color{codeframe},
  backgroundcolor=\color{codebg},
  xleftmargin=1.5em,
  framexleftmargin=1.2em,
  aboveskip=0.8em,
  belowskip=0.8em
}

\title{T2I-SEH: Towards Self-Evolving Text-to-Image Harness via \\ Effect-Conditioned Harness Learning}

\author{
Haoyuan Sun$^{1}$\thanks{Email: \texttt{haoyuan.sun@ntu.edu.sg}. $^{\dagger}$Corresponding author.} \quad
Keyan Chen$^{1}$ \quad
Yike Wu$^{1}$ \quad
Xiaoqin Zhang$^{2}$ \quad
Ling Shao$^{3}$ \quad
Shijian Lu$^{1\dagger}$ \\[4pt]
{\small $^{1}$Nanyang Technological University \quad $^{2}$Zhejiang University of Technology \quad $^{3}$University of the Chinese Academy of Sciences}}

\begin{document}
\maketitle
\input{sec/0_abstract}    
\input{sec/1_introduction}
\input{sec/2_related-work}
\input{sec/3_problem-formulation}
\input{sec/4_method}
\input{sec/5_experiments}
\input{sec/6_conclusion}

{
\small
\bibliographystyle{ieeenat_fullname}
\bibliography{main}
}

\end{document}

%% file: sec/0_abstract.tex
\begin{abstract}
Text-to-image (T2I) generation has recently been exploring a shift from conventional single-shot model invocation towards agentic systems, but its runtime control relies heavily on certain predefined strategies. The self-evolving paradigm presents a promising direction; nevertheless, existing approaches still depend on accurate state interpretation and reliable experience transfer. We propose T2I-SEH, a self-evolving T2I harness that aims to determine when prior experience can be leveraged and to what extent it could influence subsequent actions. The core design is \textbf{E}ffect-\textbf{C}onditioned \textbf{H}arness \textbf{L}earning (ECHL) which quantifies paired intervention gains, weights well-supported historical effects according to similarity and uncertainty, and yields cost-aware diagnostic beliefs. This formulation emphasizes relevant and reliable experience, while preserving prior beliefs when sufficient evidence is unavailable. ECHL then integrates cost-adjusted utility bounds with Monte Carlo estimates to determine whether to intervene, acquire additional evidence, or retain the current image. Finally, diagnostic generations refine action-specific beliefs and consolidate observed effects for future reuse. Experiments on the compositional generation task across five architecturally diverse T2I generators demonstrate that T2I-SEH enables more effective intervention selection at substantially lower action cost. In addition, when the task aligns with available interventions, resulting benefits transfer into improved generation quality effectively as reflected in GenEval 2 scores. Furthermore, evaluations on CoReBench and DPG-Bench show T2I-SEH’s robustness to the task-intervention mismatch by selectively exploiting residual cross-task utility when beneficial and otherwise defaulting to the null intervention at negligible action cost.
\end{abstract}

%% file: sec/1_introduction.tex
\section{Introduction}
\label{sec:introduction}

Recent advances in text-to-image (T2I) generation are driving a transition from conventional single-shot model invocation toward agentic systems. Rather than treating generation as an isolated forward pass, such frameworks embed the underlying generative model within a broader loop that interprets user intent, orchestrates heterogeneous models and external tools, assesses intermediate outputs, and dynamically plans subsequent actions based on the evolving execution state \citep{jiang2026genagent, he2026mind, feng2026gen, zhang2026qwen, ye2026genclaw}. Furthermore, a T2I system is characterized not only by its underlying model, but also by its harness, i.e. the surrounding system that situates the model within a broader computational workflow and governs its interaction with the environment \citep{de2026makes,202604.0428}. However, a critical bottleneck remains insufficiently explored: even when the harness is carefully designed or adapted, the runtime control strategy that determines how the model invokes is typically specified through prompts, heuristics, or fixed conventions. Accordingly, although the agent could benefit from increasingly capable external support, its decision policy rarely evolves from accumulated experience or consolidates such support into a persistent mechanism for future decision making. 

\begin{figure*}[ht]
\centering
\includegraphics[width=1.0\linewidth]{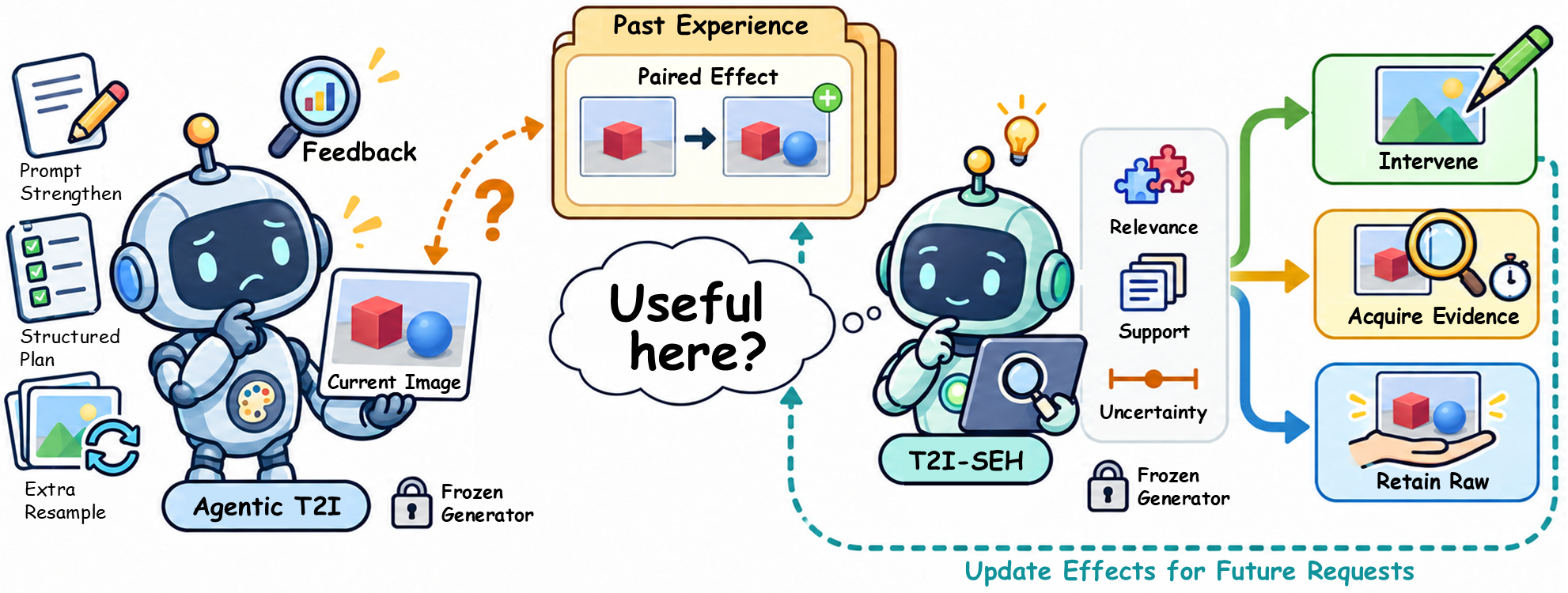}
\caption{\textbf{Overview of the T2I-SEH.} Compared with conventional agentic T2I, T2I-SEH reuses historical experience more reliably by assessing when prior experience should be trusted and how strongly it should guide future actions.}
\label{motivation}
\end{figure*}

The self-evolving paradigm offers a promising avenue for addressing this challenge. Recent agentic T2I studies have explored this paradigm through recursively generated training data \citep{zhang2025generating}, self-refined prompts \citep{wan2025maestro}, adaptive model routing \citep{jiang2026octot2i}, and the distillation of tool-mediated generation trajectories \citep{chen2026genevolve}. Motivated by the emerging successes, we further develop the T2I self-evolving harness that operationalizes self-evolution at the level of runtime control. Moreover, for LLM-based agentic systems, several recent studies have explored incorporating self-evolving paradigms into agent harness \citep{lou2026autoharness, du2026living, wang2026self, jiang2026harnessevolve}; meanwhile,  similar efforts have also emerged in agentic T2I systems, which progressively distill prior generation trajectories, execution failures, and visual feedback into reusable skills that can be leveraged in subsequent generation \citep{li2026comfyclaw,he2026gems}. Notably, these approaches generally rely on two strong assumptions: that the current generation state can be accurately interpreted, and that the retrieved experience is sufficiently relevant to be reused. Nevertheless, these assumptions remain fragile in T2I generation, where context-dependent visual feedback can lead to misinterpretation, while experience effective in one state may transfer poorly or even provide misleading guidance in another. More fundamentally, this gives rise to a central challenge: \textbf{\textit{how can a self-evolving T2I harness dynamically determine when prior experience should be trusted and how strongly it should guide future actions?}}

To this end, we introduce T2I-SEH, a self-evolving T2I harness built upon \textbf{E}ffect-\textbf{C}onditioned \textbf{H}arness \textbf{L}earning (ECHL). Given the image and context, ECHL quantifies each intervention by paired improvement effect, while retrieving active historical records that are sufficiently similar to the current context and supported by adequate observations. Then, for each action, ECHL constructs cost-aware diagnostic beliefs by weighting eligible action effects according to contextual similarity and effect uncertainty, with greater weight assigned to records exhibiting higher similarity and narrower effect intervals. The resulting mean estimates the expected benefit, and the variance quantifies uncertainty; in the absence of eligible evidence, ECHL preserves the prior belief unchanged. Furthermore, once historical or requested evidence becomes available, ECHL compares cost-adjusted utilities together with uncertainty bounds, turning beliefs into action by deciding whether the current evidence is sufficient or whether additional image generation is warranted. It retains unchanged when no corrective action exhibits plausibly positive utility, and selects the leading intervention when its lower utility bound is positive and lies within a prescribed tolerance of the upper utility bound of every competing action. If uncertainty persists, ECHL employs Monte Carlo estimation to quantify the cost-normalized value of information (VOI) of each affordable diagnostic action, terminating exploration when no action provides sufficient cost-normalized VOI and subsequently committing to the intervention with the highest posterior cost-adjusted utility. After each diagnostic generation, ECHL updates the belief associated with the tested action and progressively reduces uncertainty as evidence accumulates, while consolidating the resulting effects to support selective reuse of prior evidence in future decisions and continual evolution of the harness.

In the experiments, we employ four intervention strategies tailored to compositional generation, namely reference image retention, prompt strengthening, structured planning, and extra resampling, across five architecturally diverse T2I generators: SDXL \citep{podell2024sdxl}, SD3.5M \citep{stabilityai2024sd35medium}, FLUX.2 Klein \citep{flux-2-2025}, Z-Image-Turbo \citep{cai2025z}, and Janus-Pro-7B \citep{chen2025janus}. Firstly, we construct a controlled dataset spanning six compositional capabilities to rigorously evaluate both task-level and atomic-level correctness. Experiments demonstrate that T2I-SEH enables more effective intervention selection while incurring substantially lower action costs; moreover, results under fixed intervention strategies reveal the inherent limitations of uniform interventions, further underscoring the advantage of ECHL. Secondly, results on the GenEval 2 \citep{kamath2025geneval}, whose scenarios align with the predefined intervention settings, observed gains extend beyond observer-induced improvements in internal optimization, consistently yielding higher final-generation quality. Thirdly, experiments under task-intervention mismatch on CoReBench \citep{li2026easier} and DPG-Bench \citep{hu2024ella} demonstrate that T2I-SEH remains robust even when predefined interventions are substantially misaligned with the target task: ECHL selectively leverages the residual cross-task utility of beneficial interventions when such transferability exists, while otherwise defaulting to null intervention at negligible action cost.

In summary, our contributions are threefold:
\begin{itemize}
\item We introduce an evidence-based perspective on T2I harness self-evolution, treating prior experience as conditional support for runtime decisions and explicitly addressing when it should be trusted and how strongly it should influence subsequent actions.
\item We propose T2I-SEH with Effect-Conditioned Harness Learning, integrating paired intervention effects, uncertainty-aware beliefs, cost-adjusted utility, and value of information to guide intervention, diagnosis, or image retention across generation episodes.
\item Across five generators, experiments demonstrate improved intervention selection at substantially lower action cost, consistent gains on the GenEval 2 under task-intervention alignment, and robust performance under task-intervention mismatch through selective transfer or null intervention at negligible cost.
\end{itemize}

%% file: sec/2_related-work.tex
\section{Related Work}
\label{sec: related work}
\subsection{Agentic Text-to-Image Generation}
\label{Agentic Text-to-Image Generation}
T2I generation has increasingly shifted from reliance on a single model invocation toward an agentic inference process, in which a broader system interprets user requests, selects appropriate models or tools, evaluates intermediate outputs, and adaptively determines subsequent actions \citep{yang2024idea2img, zhao2024diffagent, wang2024genartist, jia2025chatgen, chen2025t2i}. Furthermore, recent work has substantially expanded the reasoning, state, and action spaces of such systems \citep{jiang2026genagent, he2026mind, feng2026gen, zhang2026qwen, ye2026genclaw}, suggesting that T2I generation is increasingly being characterized as a sequential decision-making process over heterogeneous operations, rather than as a static mapping. Notably, a particularly relevant direction is to make agentic T2I inference experience-aware, enabling agents to leverage prior generation attempts, tool executions, and task trajectories to adaptively select and plan heterogeneous inference-time operations \citep{jiang2026octot2i,chen2026genrouter,chen2026genevolve}. While prior work utilizes experience to adapt individual generation strategies, T2I-SEH shifts learning to the harness, asking whether a frozen model can be embedded in an experience-bearing control process that learns from its actions, reallocates future computation, and progressively improves its operational competence.

\subsection{Self-Evolving Harness in Agentic System}
\label{Self-Evolving Harness in Agentic System1}
As increasingly capable foundation models are deployed in agentic systems, adaptation is extending beyond model parameters to the persistent harness governing inference and interaction \citep{yao2023react, wu2024autogen, hong2024metagpt, yang2024swe, wang2025openhands, mei2025aios, zhang2025general, pan2026natural}. Early work showed that agents can improve through feedback, episodic memory, and reusable skills without retraining the underlying model \citep{park2023generative, madaan2023self, shinn2023reflexion, wang2024voyager, zhong2024memorybank, zhao2024expel, xu2025amem, pmlr-v267-wang25bx}. More recent studies make the harness itself an explicit target of evolution. ADAS \citep{hu2025automated} and AFlow \citep{zhang2025aflow} optimize agent architectures and executable workflows from execution feedback, while AutoHarness \citep{lou2026autoharness}, Living-Harness \citep{du2026living}, SHAPER \cite{wang2026self}, and HarnessEvolve \citep{jiang2026harnessevolve} extend this paradigm to the evolution of prompts, procedural knowledge, reusable skills, runtime code, tools, and execution logic. Collectively, these works suggest that agentic capabilities can improve through refinement of the persistent harness surrounding the model. A similar trend is emerging in agentic T2I systems: GEMS \citep{he2026gems} introduces persistent trajectory memory and reusable skills, while COMFYCLAW \citep{li2026comfyclaw} distills execution errors and visual feedback into an evolving skill library. By contrast, T2I-SEH does not expand the agent's skill set; it leverages accumulated experience to reshape the evidence underlying evaluation, action selection, and calibrated reuse of prior experience.

%% file: sec/3_problem-formulation.tex
\section{Problem Formulation}
\label{Problem Formulation}

Let $x_t\in\mathcal{X}$ denote request $t$ in an ordered stream $x_{1:T}$. A frozen generator $G_{\psi}$ is accessed through a finite intervention registry $\mathcal{A}$, which includes the \texttt{Raw} action ($a_0$) and a set of corrective actions. Fixing the interventions and costs in $\mathcal A$, the registry preserves matched randomness across diverging paths. For each request $t$, the harness maintains a persistent experience memory $\mathcal M_t$ prior to processing, while a mandatory \texttt{Raw} call returns $I_t^0$ together with the context $\kappa_t$. Executing action $a \in \mathcal A(x_t)$ produces an output $y_t(a)=G_{\psi}(x_t,a)$ and incurs cost $c(a)$, which quantifies the additional generation operations performed under the protocol, with $c(a_0)=0$ and $c(a)>0$ for all $a\in\mathcal A_+(x_t)$. Furthermore, the harness evaluates outcomes using a policy-facing function $z_{\omega}:\mathcal{X}\times\mathcal{Y}\rightarrow[0,1]$. For each $a\in\mathcal A_+(x_t)$, the paired \textbf{effect} is defined as $\delta_t(a)=z_{\omega}\left(x_t,y_t(a)\right)-z_{\omega}\left(x_t,y_t(a_0)\right)$, with corresponding contextual estimand $\theta_a(\kappa_t)=\mathbb E[\delta_t(a)|\kappa=\kappa_t]$.

Before committing to output, the harness acquires diagnostic information by selecting $a_k\in\mathcal A_+(x_t)$ at round $k$, observing $\delta_t(a_k)$, and updating the local information state $D_{t,k}$. For $K_t$ diagnostic tests, the total diagnostic cost satisfies $C^{\text{diag}}_t=\sum_{k=1}^{K_t}c(a_{t,k})\le B_t$, where $B_t$ is the hard per-request diagnostic budget. Upon committing to an intervention $\hat a_t\in\mathcal A(x_t)$, the total incremental generation cost is $C^{\text{total}}_t=C^{\text{diag}}_t+c(\hat a_t)$. We further define the cost-adjusted posterior \textbf{utility} as:
\begin{equation}
U_{t,k}(a)=\mathbb E[\theta_a(\kappa_t)\mid D_{t,k},\mathcal M_t] - \lambda c(a),
\label{utility}
\end{equation}
where $\mathcal M_t$ is the experience available before request $t$ and $\lambda\ge 0$ is fixed before evaluation. Once the decisions are finalized, the external evaluator reports the benchmark-native quality score together with the corresponding incurred cost:
\begin{equation}
Q_b=\frac{1}{|\mathcal{T}_b|}\sum_{t\in\mathcal{T}_b}q_b\left(x_t,y_t(\hat a_t)\right), C_b=\frac{1}{|\mathcal{T}_b|}\sum_{t\in\mathcal{T}_b}C^{\text{total}}_t,
\label{external-evaluation}
\end{equation}
where $\mathcal{T}_b$ is the evaluation set for benchmark $b$.

\begin{figure*}[t]
\centering
\includegraphics[width=1.0\linewidth]{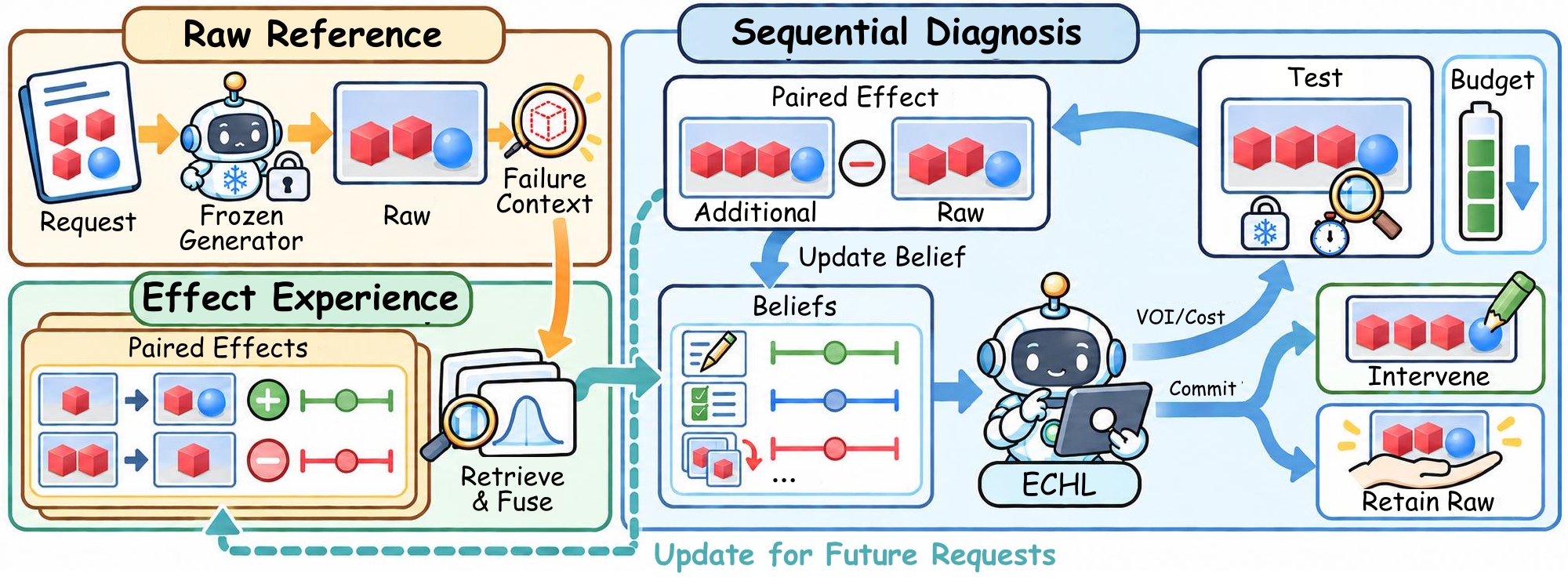}
\caption{\textbf{Overview of the ECHL.} ECHL combines effect-conditioned experience that captures paired effects with cost-aware sequential diagnosis that acquires information for cost-adjusted decision.}
\vspace{-1em}
\label{ECHL}
\end{figure*}

%% file: sec/4_method.tex
\section{Methodology}
\label{sec: methodology}

We propose T2I-SEH, a text-to-image self-evolving harness powered by Effect-Conditioned Harness Learning (ECHL), which integrates an effect-conditioned experience store and a cost-aware diagnostic strategy. Formally, we further characterize the harness dynamics as:
\begin{equation}
\begin{split}
&\mathcal H_t=(\mathcal M_t;\Omega),
\mathcal B_{t,0}=R_{\Omega}(\kappa_t,\mathcal M_t), \\
&(\tau_t,\hat a_t)=\Pi_{\Omega}(\mathcal B_{t,0},B_t),
\mathcal M_{t+1}=F_{\Omega}(\mathcal M_t;e_t),
\end{split}
\end{equation}
where $R_{\Omega}$ initializes effect beliefs from retrieved experience, $\Pi_{\Omega}$ generates diagnostic trace $\tau_t$ and committed action $\hat a_t$, and $F_{\Omega}$ updates the persistent memory with harness-visible tested events; the specification $\Omega$ fixes the generator, intervention registry, update rule, and decision thresholds.

\subsection{Effect-Conditioned Experience}
\label{Effect-Conditioned Experience}

A self-evolving T2I harness requires an experience representation that disentangles intervention efficacy from prompt difficulty, since absolute image scores may obscure this distinction: easy prompts can succeed irrespective of the applied intervention, whereas difficult ones may remain imperfect despite an effective correction. ECHL therefore encodes experience as paired improvement over the mandatory reference image under registered coupling.

For $n$ tested outcomes sharing the same context–action key, the memory update aggregates their paired effects through the empirical mean $\bar{\delta}_n =1/n\cdot\sum_{j=1}^{n}\delta_j$ and the corresponding empirical variance $\widehat{\sigma}_n^2 =1/(n-1)\cdot\sum_{j=1}^{n}(\delta_j-\bar{\delta}_n)^2$,
which serves as an estimate of the current uncertainty associated with the effect. Under a Gaussian approximation, we define the 95\% confidence interval as $[\ell_n,u_n]=[\bar{\delta}_n-h_n,\bar{\delta}_n+h_n]$, with $h_n=1.96\sqrt{\max(\widehat{\sigma}_n^2,0.25)/n}$; the variance floor is introduced to prevent the interval from becoming spuriously narrow when the observed variability is small. In addition, a support-dependent confidence score is defined as $\chi_n=n/(n+2)$, assigning greater weight to context-action estimates supported by more observations.

Furthermore, given the observed failure context $\kappa_t$, relevant historical experience is retrieved by evaluating the structured similarity between $\kappa_t$ and each stored context $\kappa_i$:
\begin{equation}
\begin{split}
\operatorname{sim}(\kappa_t,\kappa_i)=
\lambda_{\text{failure}} s_{\text{failure}}&+\lambda_{\text{generator}} s_{\text{generator}} \\ + \lambda_{\text{severity}} s_{\text{severity}}&+\lambda_{\text{context}} s_{\text{context}}.
\label{similarity}
\end{split}
\end{equation}
Accordingly, only active memory records with sufficient contextual similarity and empirical support are retained for retrieval.
For request $t$, the resulting admissible experience set is defined as: 
\begin{equation}
\begin{split}
\mathcal{R}_t=\{&m_i\in\mathcal M_t\; | \; \\ \rho_i&=\textsc{Active},\operatorname{sim}(\kappa_t,\kappa_i)\ge0.75,\ 
n_i\ge10\}.
\end{split}
\label{eq:retrieval-criteria}
\end{equation}
Let $\mathcal R_{t,a}=\{m_i\in\mathcal R_t \mid a_i=a\}$
denote this action-specific evidence set. For each $m_i\in\mathcal R_{t,a}$, we estimate relevance $w_i$ and uncertainty $v_i^M$ from the width of its effect interval:
\begin{equation}
\begin{split}
w_i&=\max\left\{
\operatorname{sim}(\kappa_t,\kappa_i)\chi_i,10^{-6}\right\}, 
\\
v_i^M&=\max\left\{
\left(\frac{u_i-\ell_i}{2\times1.96}\right)^2, 10^{-4}\right\}.
\end{split}
\end{equation}
Hence, records receive greater influence on the updated belief when they are more similar to the current context, have higher confidence, or provide more precise effect estimate. Finally, the updated effect belief is obtained by integrating the frozen contextual prior with the retrieved historical evidence in a precision-weighted manner, providing the basis for subsequent action selection: 
\begin{equation} 
\begin{split}
\mu_a&=\bigg(\frac{\mu_a^0}{v_a^0}+\sum_{i} \frac{w_i}{v_i^M}\bar{\delta}_i\bigg)\Big/\bigg(\frac{1}{v_a^0}+\sum_{i}\frac{w_i}{v_i^M}\bigg),
\\
v_a&=\bigg(\frac{1}{v_a^0}+\sum_{i}\frac{w_i}{v_i^M}\bigg)^{-1}.
\end{split}
\label{memory-fusion}
\end{equation}
The fused belief influences downstream decision-making through its mean and variance. Specifically, $\mu_a$ characterizes the expected effect of intervention $a$ and thereby influences the ranking, while $v_a$ quantifies the residual uncertainty and determines whether further diagnostic image generation is warranted and when it should terminate. In particular, when no action-specific historical evidence is available, i.e., $\mathcal{R}_{t,a}=\varnothing$, ECHL retains the original prior, yielding $(\mu_a,v_a)=(\mu_a^0,v_a^0)$.

\subsection{Cost-Aware Sequential Diagnosis}
\label{Cost-Aware Sequential Diagnosis}

Historical experience provides valuable guidance for future T2I decision-making, yet its reliability is inherently context-dependent, while exhaustive evaluation incurs substantial costs by requiring the generation of additional candidates. Accordingly, ECHL models each diagnostic generation as an information-acquisition step whose utility derives from reducing uncertainty sufficiently to alter the final cost-adjusted decision. Let $\mathcal A(x_t)\subseteq\mathcal A$ denote the feasible action subset for $x_t$, and define that $\mathcal A_+(x_t)=\mathcal A(x_t)\setminus\{a_0\}$. At diagnostic round $k$, ECHL summarizes the evidence available for request $t$, including the persistent memory $\mathcal{M}_t$ and the information after $k$ diagnostic steps $D_{t,k}$, as a factorized Gaussian belief state $\mathcal{B}_{t,k}=\left\{(\mu_{t,k,a},v_{t,k,a})\right\}_{a\in\mathcal{A}_+(x_t)}$, where the effect of each feasible corrective action \(a\) is approximated as $\theta_a(\kappa_t)\approx\mathcal{N}(\mu_{t,k,a},v_{t,k,a})$. Combining with \Cref{utility}, the cost-adjusted utility of each corrective action is given by:
\begin{equation}
U_{t,k}(a)=\mu_{t,k,a}-\lambda c(a).
\end{equation}
Accordingly, the lower and upper bounds on the cost-adjusted utility are defined as:
\begin{equation}
\begin{split}
\underline{U}_{t,k}(a)=\mu_{t,k,a}-\lambda c(a)-1.96\sqrt{v_{t,k,a}},
\\
\overline{U}_{t,k}(a)=\mu_{t,k,a}-\lambda c(a)+1.96\sqrt{v_{t,k,a}}.
\end{split}
\end{equation}

When an action $a$ is diagnostically tested, the observed paired effect $d_{t,k+1}=\delta_t(a)$ is modeled as a measurement of the action’s latent effectiveness. The corresponding belief is then updated by integrating the new observation with the evidence accumulated prior to the diagnostic test:
\begin{equation}
\begin{split}
\mu_{t,k+1,a}&=v_{t,k+1,a}\left(\frac{\mu_{t,k,a}}{v_{t,k,a}}+\frac{d_{t,k+1}}{v_{\text{obs}}}\right),
\\
v_{t,k+1,a}&=\left(v_{t,k,a}^{-1}+v_{\text{obs}}^{-1}\right)^{-1}.
\end{split}
\label{belief-update}
\end{equation}
This precision-weighted update integrates the posterior mean $\mu_{t,k,a}$ with the observed effect $d_{t,k+1}$, while reducing the associated uncertainty from $v_{t,k,a}$ to $v_{t,k+1,a}$. Then, the updated beliefs are subsequently used to assess whether additional diagnostic generation is needed to improve the decision. Moreover, action with the highest posterior utility is defined as \(a^\mu_{t,k}=\arg\max_{a\in\mathcal A(x_t)}U_{t,k}(a)\), with ties resolved in favor of \texttt{Raw}. The residual optimistic decision loss is then quantified as:
\begin{equation}
\mathcal{L}(\mathcal B_{t,k})=\left[\max_{a\in\mathcal A(x_t)}\bigg(\overline{U}_{t,k}(a)-U_{t,k}(a^\mu_{t,k})\bigg)\right]_+,
\label{unresolved-loss}
\end{equation}
where $[x]_+=\max(x,0)$. This quantity measures the gap between the most favorable utility under current uncertainty and the posterior mean utility of the selected action, with larger values indicating greater potential benefit from additional evidence.

Furthermore, the potential decision benefit of an additional diagnostic observation is estimated via Monte Carlo approximation under the current belief state. For each affordable corrective action $a$, $M$ candidate outcomes are sampled from the predictive distribution $\mathcal{N}\left(\mu_a,v_a+v_{\text{obs}}\right)$, and each sample $m$ is propagated through \Cref{belief-update} to construct the corresponding belief state $\mathcal B^{a,m}$. The expected decision benefit of testing action $a$ is quantified via Monte Carlo approximation to the value of information (VOI), defined as the expected reduction in decision loss of the test:
\begin{equation}
\widehat{\operatorname{VOI}}(a)=\left[\mathcal L(\mathcal B)-\frac{1}{M}\sum_{m=1}^{M}\mathcal L\left(\mathcal B^{a,m}\right)\right]_+.
\label{voi}
\end{equation}
Here, $\mathcal L(\mathcal B)$ is the current optimistic decision loss, while the second term approximates its expected value after testing action $a$. Moreover, to account for heterogeneous generation costs, each affordable action is ranked by its estimated value of information per unit cost, $\widehat{\operatorname{VOI}}(a)/c(a)$, and the highest-ranked action is selected for the next diagnostic test.

ECHL determines whether to continue diagnostic testing through following checks. First, it evaluates whether any corrective action can plausibly outperform the zero-utility \texttt{Raw} baseline; if $\max_{a\in\mathcal A_+(x_t)}\overline U_{t,k}(a)<0$, no corrective action remains favorable even under its optimistic utility bound, and \texttt{Raw} is therefore retained. Second, early commitment is permitted when the leading corrective action is sufficiently supported by the current belief state: ECHL commits to the highest-mean-utility action $a^\mu$ only if its lower utility bound is positive and remains within $\epsilon$ of the strongest competing candidate's upper bound, i.e. $\underline {U}_{t,k}{(a^\mu)}>0$ and $ \underline {U}_{t,k}{(a^\mu)}\ge\max_{b\in\mathcal A(x_t)\setminus\{a^\mu\}}\overline {U}_{t,k}(b)-\epsilon$. Third, if the preferred action remains uncertain, an additional affordable diagnostic test is performed when $\max_a \left[\widehat{\operatorname{VOI}}(a)/c(a)\right]>\eta$, ensuring that additional tests are conducted only when their expected decision VOI is sufficiently high relative to the cost. Finally, when no informative or affordable diagnostic test remains, ECHL terminates the diagnostic process and selects the action with the highest current utility $\hat a_t=\arg\max_{a\in\mathcal A(x_t)}U_{t,k}(a)$.

\begin{table*}[t]
\centering
\caption{Task Success Rate (TSR), Constraint Satisfaction Rate (CSR), and mean action cost evaluated by Qwen3.8-27B on the controlled dataset, using 360 instances for each generator under budgets $B\in\{1,2,4\}$. Raw denotes the initial image generated without intervention. PS, SP, and ER denote Prompt Strengthening, Structured Planning, and Extra Resampling, respectively.}
\label{Q1}
\setlength{\tabcolsep}{5pt}
\small
\vspace{-1em}
\begin{tabular}{@{}lc*{5}{c}|@{\hspace{10pt}}*{5}{c}|@{\hspace{10pt}}*{2}{c}@{}}
\toprule
\multirow{2}{*}{Generator}
& \multirow{2}{*}{$B$}
& \multicolumn{5}{c}{TSR (\%) $\uparrow$}
& \multicolumn{5}{c}{CSR (\%) $\uparrow$}
& \multicolumn{2}{c}{Action Cost $\downarrow$} \\
\cmidrule{3-7}\cmidrule{8-12}\cmidrule{13-14}
& & Raw & PS & SP & ER & \textbf{ECHL}
  & Raw & PS & SP & ER & \textbf{ECHL}
  & ER & \textbf{ECHL} \\
\midrule

\multirow{3}{*}{SDXL}
& 1 & 23.06 & 26.11 & 20.00 & - & \textbf{26.94}
    & 36.08 & 37.25 & 30.50 & - & \textbf{39.56}
    & - & \textbf{0.15} \\
& 2 & 23.06 & 26.11 & 20.00 & 28.06 & \textbf{28.33}
    & 36.08 & 37.25 & 30.50 & 40.39 & \textbf{41.08}
    & 2.00 & \textbf{0.21} \\
& 4 & 23.06 & 26.11 & 20.00 & 28.06 & \textbf{30.83}
    & 36.08 & 37.25 & 30.50 & 40.39 & \textbf{43.42}
    & 2.00 & \textbf{0.35} \\
\midrule

\multirow{3}{*}{SD3.5M}
& 1 & 54.72 & 46.94 & 46.39 & - & \textbf{55.56}
    & 70.61 & 63.36 & 62.58 & - & \textbf{71.22}
    & - & \textbf{0.10} \\
& 2 & 54.72 & 46.94 & 46.39 & \textbf{60.00} & 58.33
    & 70.61 & 63.36 & 62.58 & \textbf{75.72} & 73.28
    & 2.00 & \textbf{0.16} \\
& 4 & 54.72 & 46.94 & 46.39 & 60.00 & \textbf{60.56}
    & 70.61 & 63.36 & 62.58 & 75.72 & \textbf{76.44}
    & 2.00 & \textbf{0.25} \\
\midrule

\multirow{3}{*}{FLUX.2 Klein}
& 1 & 56.67 & 56.94 & 57.22 & - & \textbf{58.06}
    & 68.25 & 66.61 & 68.42 & - & \textbf{68.75}
    & - & \textbf{0.06} \\
& 2 & 56.67 & 56.94 & 57.22 & 60.83 & \textbf{61.11} & 68.25 & 66.61 & 68.42 & 70.28 & \textbf{71.19}
    & 2.00 & \textbf{0.11} \\
& 4 & 56.67 & 56.94 & 57.22 & 60.83 & \textbf{62.22}
    & 68.25 & 66.61 & 68.42 & 70.28 & \textbf{71.94}
    & 2.00 & \textbf{0.15} \\
\midrule

\multirow{3}{*}{Z-Image-Turbo}
& 1 & 70.56 & 70.28 & 61.67 & - & \textbf{72.22}
    & 78.44 & 79.06 & 72.28 & - & \textbf{79.81}
    & - & \textbf{0.05} \\
& 2 & 70.56 & 70.28 & 61.67 & 73.33 & \textbf{74.17}
    & 78.44 & 79.06 & 72.28 & 81.06 & \textbf{81.89}
    & 2.00 & \textbf{0.11} \\
& 4 & 70.56 & 70.28 & 61.67 & 73.33 & \textbf{75.00}
    & 78.44 & 79.06 & 72.28 & 81.06 & \textbf{82.72}
    & 2.00 & \textbf{0.13} \\
\midrule

\multirow{3}{*}{Janus-Pro-7B}
& 1 & 46.11 & 39.44 & 5.28 & - & \textbf{49.17}
    & 60.33 & 49.42 & 15.42 & - & \textbf{63.25}
    & - & \textbf{0.11} \\
& 2 & 46.11 & 39.44 & 5.28 & \textbf{50.83} & 49.17
    & 60.33 & 49.42 & 15.42 & \textbf{63.75} & 64.19
    & 2.00 & \textbf{0.11} \\
& 4 & 46.11 & 39.44 & 5.28 & 50.83 & \textbf{51.39}
    & 60.33 & 49.42 & 15.42 & 63.75 & \textbf{65.53}
    & 2.00 & \textbf{0.26} \\
\bottomrule
\end{tabular}
\label{tab:q1}
\vspace{-1em}
\end{table*}

Additionally, the persistent memory enables continual adaptation across requests. Once ECHL commits to an action for request $t$, the paired effects observed during diagnosis are consolidated into experience records and stored in $\mathcal M_{t+1}$, so that the resulting evidence is available to subsequent decisions. When retrieved for a subsequent request, the relevant experience record is utilized to update both expected utility and associated uncertainty of the intervention. The algorithmic procedure is illustrated in \Cref{ECHL}, and the corresponding pseudocode is provided in the Appendix.

%% file: sec/5_experiments.tex
\section{Experiments}
\label{sec: experiments}

\subsection{Experimental Setup}
\label{sec:exp-setup}

\textbf{Tasks and Interventions. } For simplicity, we apply the compositional generation setting in which the generated outputs can be readily verified. For each request, a shared predefined action set comprising four types of interventions is considered: retaining the reference image (\textbf{\texttt{Raw}}), Prompt Strengthening (\textbf{\texttt{PS}}), Structured Planning (\textbf{\texttt{SP}}), and Extra Resample (\textbf{\texttt{ER}}). Specifically, we further assign cost to each intervention to reflect the additional overhead: \textbf{\texttt{Raw}} leaves the input unchanged and incurs zero additional cost; \textbf{\texttt{PS}} preserves the original task structure while explicitly strengthening requirements that are prone to failure, incurring an additional action cost of 1; \textbf{\texttt{SP}} first decomposes a complex request into structured constraints and then constructs the generation instructions accordingly, incurring an additional action cost of 1; and \textbf{\texttt{ER}} leaves the prompt unchanged but introduces additional sampling opportunity, yielding two candidates in total, where the final output is selected by an internal selector, resulting in an additional action cost of 2. In our experiments, we set budget to $B\in\{1,2,4\}$ and evaluate performance under different budget levels. Results are averaged over five random seeds. In addition, the Appendix also reports application of ECHL to image editing tasks.

\noindent\textbf{Research Questions.} First, under a fixed generator, intervention set, and scoring rule, we investigate whether T2I-SEH can select more appropriate interventions with lower cost (\textbf{RQ1}). Second, we aim to examine whether the observed gains arise merely from improved optimization against the internal observer, or instead transfer into genuine improvements in final generation quality on compositional-generation benchmarks (\textbf{RQ2}). Finally, we investigate whether T2I-SEH can selectively exploit transferable interventions and otherwise abstain under task--intervention mismatch, while preserving the underlying generator's performance (\textbf{RQ3}).

\begin{table*}[t]
\centering
\caption{Internal Success Rate (ISR), GenEval 2 Soft-TIFA, and mean action cost evaluated on the GenEval 2 compositional-generation benchmark under budgets $B\in\{1,2,4\}$. Raw denotes the initial image generated without intervention. PS, SP, and ER denote Prompt Strengthening, Structured Planning, and Extra Resampling, respectively.}
\label{Q2}
\small
\vspace{-1em}
\setlength{\tabcolsep}{5pt}
\begin{tabular}{@{}lc*{5}{c}|@{\hspace{10pt}}*{5}{c}|@{\hspace{10pt}}*{2}{c}@{}}
\toprule
\multirow{2}{*}{Generator}
& \multirow{2}{*}{$B$}
& \multicolumn{5}{c}{ISR (\%) $\uparrow$}
& \multicolumn{5}{c}{GenEval 2 Soft-TIFA $\uparrow$}
& \multicolumn{2}{c}{Action Cost $\downarrow$} \\
\cmidrule{3-7}\cmidrule{8-12}\cmidrule{13-14}
& & Raw & PS & SP & ER & \textbf{ECHL}
  & Raw & PS & SP & ER & \textbf{ECHL}
  & ER & \textbf{ECHL} \\
\midrule

\multirow{3}{*}{SDXL}
& 1
& 61.15 & \textbf{83.48} & 51.08 & - & 75.95
& 7.97 & 6.93 & 5.35 & - & \textbf{8.38}
& - & \textbf{0.15} \\
& 2
& 61.15 & 83.48 & 51.08 & 76.50 & \textbf{89.03}
& 7.97 & 6.93 & 5.35 & 8.15 & \textbf{8.59}
& 2.00 & \textbf{0.28} \\
& 4
& 61.15 & 83.48 & 51.08 & 76.50 & \textbf{91.95}
& 7.97 & 6.93 & 5.35 & 8.15 & \textbf{8.73}
& 2.00 & \textbf{0.34} \\

\midrule

\multirow{3}{*}{SD3.5M}
& 1
& 79.98 & 90.43 & 72.35 & - & \textbf{92.10}
& 19.04 & 16.31 & 11.53 & - & \textbf{19.73}
& - & \textbf{0.12} \\
& 2
& 79.98 & 90.43 & 72.35 & 91.25 & \textbf{96.20}
& 19.04 & 16.31 & 11.53 & \textbf{20.22} & 19.92
& 2.00 & \textbf{0.16} \\
& 4
& 79.98 & 90.43 & 72.35 & 91.25 & \textbf{97.80}
& 19.04 & 16.31 & 11.53 & \textbf{20.22} & 20.06
& 2.00 & \textbf{0.19} \\

\midrule

\multirow{3}{*}{FLUX.2 Klein}
& 1
& 89.55 & 97.35 & 95.28 & - & \textbf{98.83}
& 25.48 & 24.77 & 23.51 & - & \textbf{26.78}
& - & \textbf{0.09} \\
& 2
& 89.55 & 97.35 & 95.28 & 94.98 & \textbf{99.63}
& 25.48 & 24.77 & 23.51 & 26.14 & \textbf{26.84}
& 2.00 & \textbf{0.10} \\
& 4
& 89.55 & 97.35 & 95.28 & 94.98 & \textbf{99.70}
& 25.48 & 24.77 & 23.51 & 26.14 & \textbf{26.84}
& 2.00 & \textbf{0.10} \\

\midrule

\multirow{3}{*}{Z-Image-Turbo}
& 1
& 94.83 & 98.45 & 97.23 & - & \textbf{99.05}
& 31.50 & 28.74 & 28.42 & - & \textbf{32.26}
& - & \textbf{0.04} \\
& 2
& 94.83 & 98.45 & 97.23 & 97.13 & \textbf{99.83}
& 31.50 & 28.74 & 28.42 & 31.76 & \textbf{32.27}
& 2.00 & \textbf{0.05} \\
& 4
& 94.83 & 98.45 & 97.23 & 97.13 & \textbf{99.90}
& 31.50 & 28.74 & 28.42 & 31.76 & \textbf{32.27}
& 2.00 & \textbf{0.05} \\

\midrule

\multirow{3}{*}{Janus-Pro-7B}
& 1
& 67.48 & 80.25 & 34.05 & - & \textbf{86.10}
& 13.26 & 11.68 & 5.19 & - & \textbf{13.73}
& - & \textbf{0.19} \\
& 2
& 67.48 & 80.25 & 34.05 & 78.85 & \textbf{89.03}
& 13.26 & 11.68 & 5.19 & 14.24 & \textbf{14.35}
& 2.00 & \textbf{0.22} \\
& 4
& 67.48 & 80.25 & 34.05 & 78.85 & \textbf{90.98}
& 13.26 & 11.68 & 5.19 & 14.24 & \textbf{14.47}
& 2.00 & \textbf{0.25} \\
\bottomrule
\end{tabular}
\end{table*}

\noindent\textbf{Models and Datasets.} To capture architectural diversity across modern generative models, we evaluate five representative generators. Specifically, Stable Diffusion XL (SDXL) \citep{podell2024sdxl}, Stable Diffusion 3.5 Medium (SD3.5M) \citep{stabilityai2024sd35medium}, FLUX.2 Klein \citep{flux-2-2025}, and Z-Image-Turbo \citep{cai2025z} instantiate diffusion-based or flow-matching-based generation, while Janus-Pro-7B \citep{chen2025janus} additionally represents the autoregressive paradigm. To address RQ1, we construct a controlled dataset comprising 360 test instances based on predefined templates, structured targets, and atomic evaluation constraints, with additional details provided in the Appendix. Candidate outputs are evaluated using Qwen3.8-27B \citep{yang2025qwen3} as the atomic evaluator. To address RQ2, we conduct evaluation on the GenEval 2 benchmark \citep{kamath2025geneval}, adopting the Qwen3-VL-8B-Instruct-based Soft-TIFA implementation \citep{bai2025qwen3}. To address RQ3, we evaluate T2I-SEH under task--intervention mismatch on CoReBench \citep{li2026easier} and DPG-Bench \citep{hu2024ella}, which respectively target reasoning and long-prompt following, to assess whether ECHL can selectively exploit transferable interventions when beneficial while otherwise abstaining with low overhead. Notably, during evaluation, T2I-SEH has access only to the frozen generic atomic signal provided by observer Qwen2.5-VL-3B \citep{bai2025qwen25vltechnicalreport}, while official benchmark scores remain strictly hidden from system.

\subsection{Better Selection at Lower Cost}
\label{sec:q1}

\begin{figure}[t]
\centering
\includegraphics[width=1.0\linewidth]{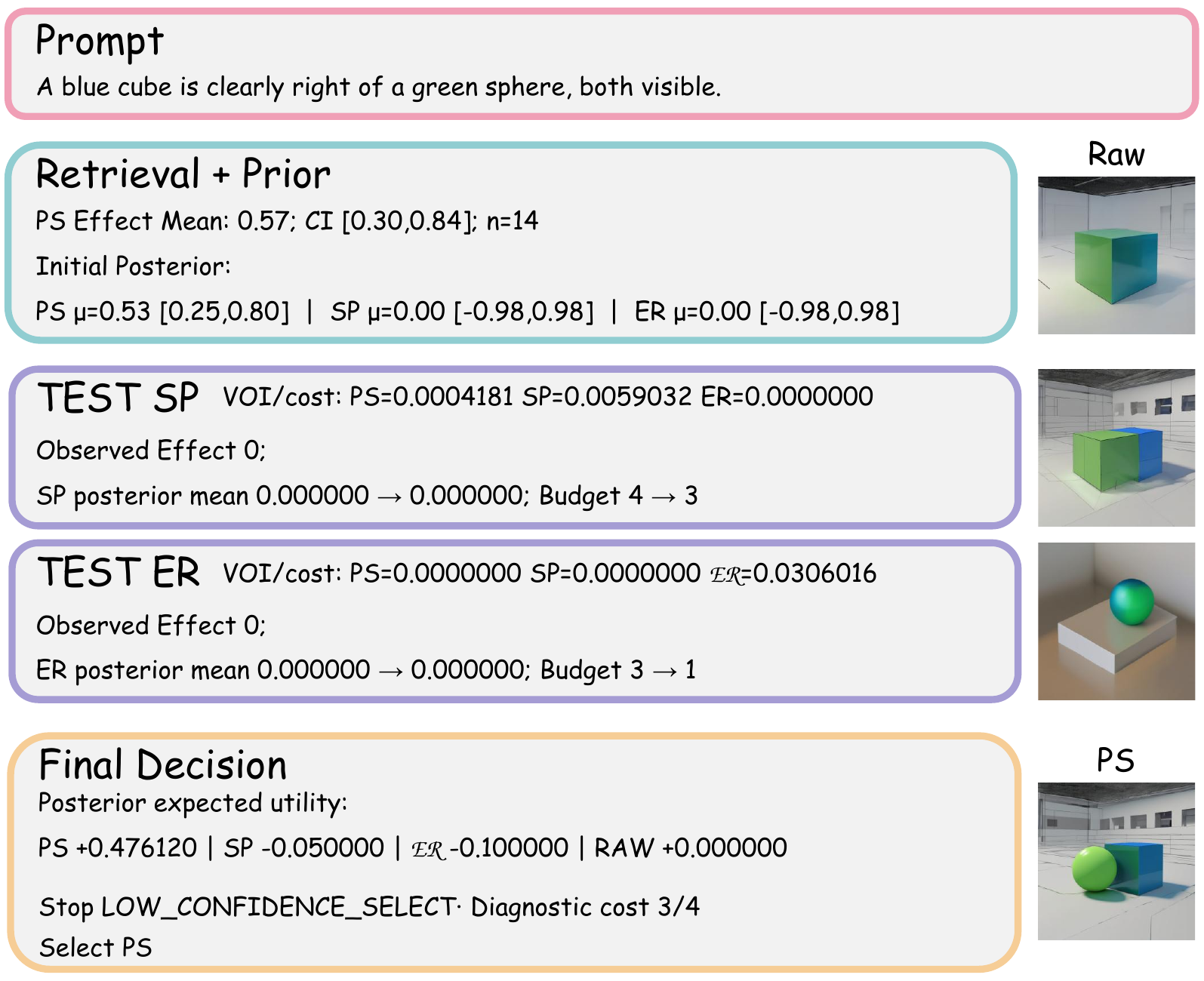}
\vspace{-1em}
\caption{ECHL decision process with retrieved memory.}
\label{example1}
\end{figure}

\begin{figure}[t]
\centering
\includegraphics[width=1.0\linewidth]{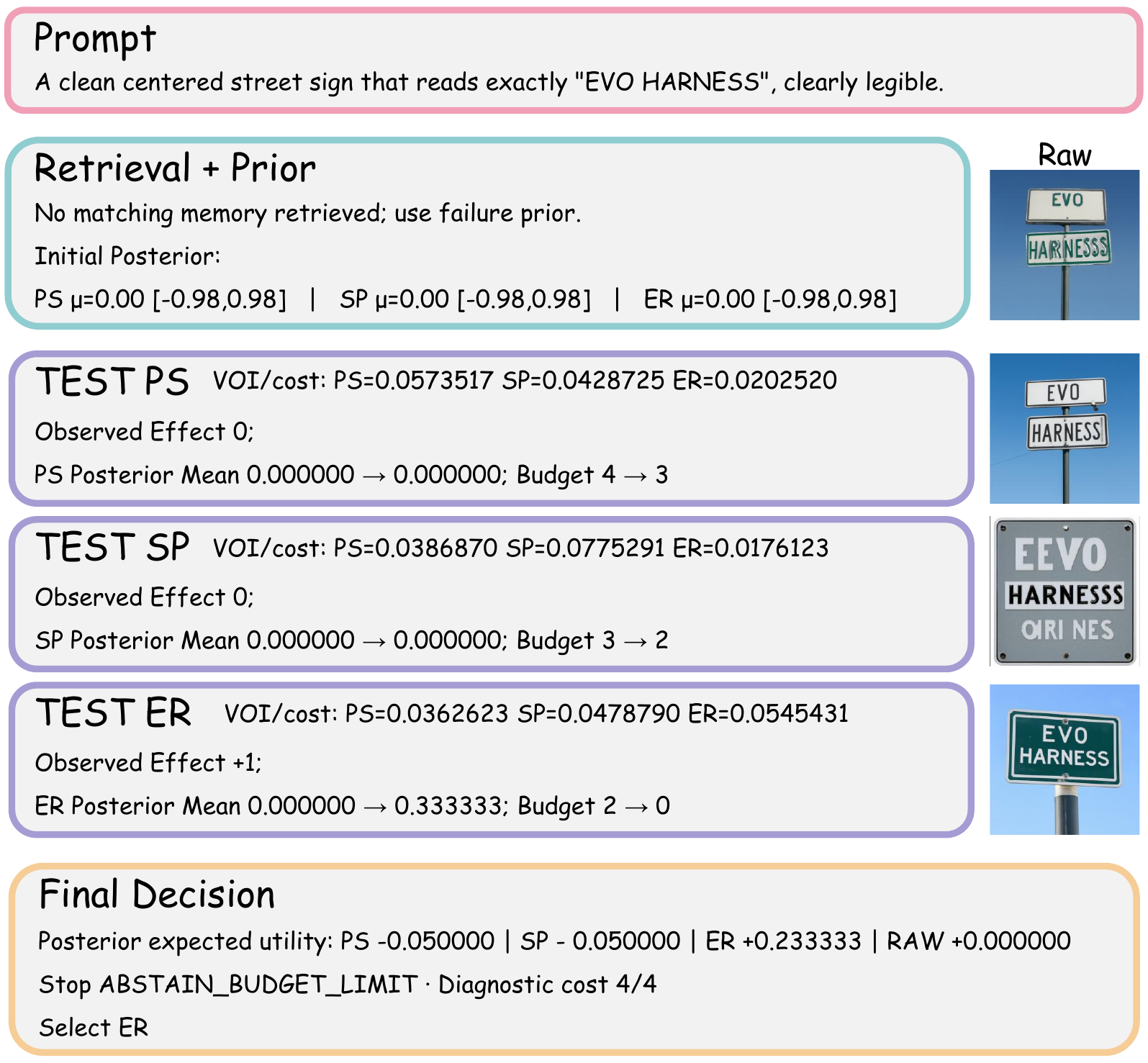}
\vspace{-1em}
\caption{ECHL decision process without matching memory.}
\vspace{-1em}
\label{example2}
\end{figure}

In this section, we give an answer to RQ1: T2I-SEH selects more appropriate interventions while incurring lower action cost. First, we define Task Success Rate (TSR) as the proportion of tasks that satisfy all registered atomic constraints, thereby capturing complete task-level correctness, while Constraint Satisfaction Rate (CSR) measures partial correctness by averaging the proportion of satisfied atomic constraints across tasks. As shown in \Cref{tab:q1}, ECHL consistently achieves a favorable trade-off between output quality and action cost across all five generators. Under the $B=4$ setting, ECHL establishes a consistent advantage in both TSR and CSR over all baselines across the five generators. Additionally, despite ER being the strongest fixed-intervention baseline, ECHL surpasses it across all five generators at $B=4$, while requiring only a fraction of ER's action cost. Taken together, these results underscore the importance of conditioned intervention routing, rather than relying on a universally beneficial prompt strengthening or indiscriminate resampling strategy. Additionally, Figures \ref{example1} and \ref{example2} further illustrate ECHL's self-evolving decision-making process. Figure \ref{example1} traces its decisions when retrieved experience informs its initial beliefs, and Figure \ref{example2} traces its decisions when memory retrieval has no match.

\subsection{Internal Gains Transfer to Generation Quality}
\label{sec:q2}

In this section, we give an answer to RQ2: the observed gains extend beyond improvement induced by the internal observer, manifesting as substantive enhancements in the final generation quality that are consistently captured by the official compositional-generation benchmark score. As shown in \Cref{Q2}, ECHL improves the official GenEval 2 over the corresponding Raw baseline across all generators and evaluated budget settings, demonstrating that its gains extend beyond the internal evaluation signal. Moreover, it is worth noting that the rankings induced by the internal and external metrics are not fully consistent. For instance, on SDXL, PS substantially increases ISR from 61.15\% to 83.48\% while reducing the GenEval 2 official score from 7.97 to 6.93. Taken together, these results indicate that the official GenEval 2 score does not merely reflect the preferences of the internal observer; rather, the internal signal often serves as an informative yet imperfect proxy for external generation quality, which ECHL leverages to achieve consistent gains over Raw despite generator-dependent variation in intervention effectiveness and gain magnitude.

\begin{table}[t]
\centering
\caption{Performance under task-intervention mismatch, evaluated on CoReBench and DPG-Bench, together with the mean action cost of ECHL under budgets $B\in\{1,2,4\}$. Raw denotes initial image without intervention, and ER stands for Extra Resampling.}
\vspace{-1em}
\label{Q3}
\setlength{\tabcolsep}{4pt}
\small
\begin{tabular}{@{}lc ccc|ccc@{}}
\toprule
\multirow{2}{*}{Generator}
& \multirow{2}{*}{$B$}
& \multicolumn{3}{c|}{CoReBench}
& \multicolumn{3}{c}{DPG-Bench} \\
\cmidrule(lr){3-5}
\cmidrule(lr){6-8}
& & Raw & ECHL & \makecell{Action \\ Cost}
& Raw & ECHL & \makecell{Action \\ Cost} \\
\midrule
\multirow{3}{*}{SDXL}
& 1
& 66.45 & \textbf{66.59} & 0.029
& 73.25 & \underline{73.26} & 0.009 \\
& 2
& 66.45 & \textbf{66.60} & 0.042
& 73.25 & \underline{73.28} & 0.011 \\
& 4
& 66.45 & \textbf{66.62} & 0.058
& 73.25 & \underline{73.30} & 0.014 \\
\midrule
\multirow{3}{*}{SD3.5M}
& 1
& 70.10 & \textbf{70.17} & 0.015
& 84.18 & \underline{84.18} & 0.002 \\
& 2
& 70.10 & \textbf{70.18} & 0.022
& 84.18 & \underline{84.18} & 0.002 \\
& 4
& 70.10 & \textbf{70.21} & 0.033
& 84.18 & \underline{84.18} & 0.003 \\
\midrule
\multirow{3}{*}{\makecell{FLUX.2\\Klein}}
& 1
& 76.19 & \textbf{76.24} & 0.011
& 85.67 & \underline{85.67} & 0.003 \\
& 2
& 76.19 & \textbf{76.24} & 0.014
& 85.67 & \underline{85.67} & 0.003 \\
& 4
& 76.19 & \textbf{76.25} & 0.018
& 85.67 & \underline{85.67} & 0.003 \\
\midrule
\multirow{3}{*}{\makecell{Z-Image\\-Turbo}}
& 1
& 75.74 & \textbf{75.78} & 0.011
& 84.87 & \underline{84.87} & 0.005 \\
& 2
& 75.74 & \textbf{75.79} & 0.015
& 84.87 & \underline{84.87} & 0.005 \\
& 4
& 75.74 & \textbf{75.80} & 0.018
& 84.87 & \underline{84.87} & 0.005 \\
\midrule
\multirow{3}{*}{\makecell{Janus-Pro\\-7B}}
& 1
& 66.45 & \textbf{66.60} & 0.022
& 82.97 & \underline{83.00} & 0.002 \\
& 2
& 66.45 & \textbf{66.68} & 0.040
& 82.97 & \underline{83.00} & 0.002 \\
& 4
& 66.45 & \textbf{66.76} & 0.054
& 82.97 & \underline{83.00} & 0.003 \\
\bottomrule
\end{tabular}
\end{table}

\subsection{Robustness under Task-Intervention Mismatch}
In this section, we give an answer to RQ3: T2I-SEH remains robust under task-intervention mismatch by leveraging ECHL to selectively exploit residual cross-task utility when beneficial and otherwise abstain at negligible cost, thereby preserving the performance of the underlying generator. By construction, both CoReBench and DPG-Bench instantiate task-intervention mismatch settings, where the intervention operators developed for compositional generation are not explicitly designed to address reasoning  or long-prompt tasks. On CoReBench, ECHL surpasses Raw across all configurations, delivering an average improvement of 0.11 points and a maximum gain of 0.31 points, while incurring mean action costs of only 0.011-0.058, indicating that ECHL can identify and exploit the sparse subset of interventions that remains transferable.  Moreover, DPG-Bench exhibits substantially less transferable utility, leading ECHL to predominantly select the null intervention and incur near-zero mean action costs of 0.002-0.014; correspondingly, performance remains highly stable relative to Raw. Taken together, these results show that ECHL adapts its intervention strategy to the degree of cross-task transferability, invoking beneficial interventions when transferable utility is present and otherwise abstaining to preserve the underlying generator's performance.

%% file: sec/6_conclusion.tex
\section{Conclusion}
\label{sec: conclusion}
In this work, we introduce T2I-SEH, a self-evolving T2I harness that determines when prior experience should be trusted and how strongly it should influence subsequent actions. Through the Effect-Conditioned Harness Learning (ECHL), paired intervention effects are weighted by contextual relevance and uncertainty, then combined with cost-aware evidence acquisition to guide intervention, diagnosis, or image retention. Experiments across five generators demonstrate that T2I-SEH enables more effective intervention selection and balances generation quality with action cost, yielding consistent quality gains when tasks align with the intervention set. Under task-intervention mismatch, evaluations on CoReBench and DPG-Bench show that it selectively applies transferable interventions and otherwise retains the original image at negligible cost. Future work could further extend the methodology to more complex generation and design scenarios, where reliable cross-stage experience reuse may offer additional benefits.